\documentclass[sigconf, nonacm
]{acmart}
\usepackage{amsmath}
\usepackage{tikz}
\usetikzlibrary{positioning, shapes.geometric, arrows.meta}
\DeclareMathOperator*{\argmax}{arg\,max}
\AtBeginDocument{%
  }

\setcopyright{acmlicensed}
\copyrightyear{2018}
\acmYear{2018}
\acmDOI{XXXXXXX.XXXXXXX}
\acmConference[Conference acronym 'XX]{Make sure to enter the correct
  conference title from your rights confirmation email}{June 03--05,
  2018}{Woodstock, NY}
\acmISBN{978-1-4503-XXXX-X/2018/06}

\begin{document}

\title[Taxonomy-Aligned Risk Extraction from 10-K Filings]{Taxonomy-Aligned Risk Extraction from 10-K Filings with Autonomous Improvement Using LLMs}

\author{Rian Dolphin}
    \authornote{Both authors contributed equally to this research. Contact at \href{mailto:research@massive.com}{research@massive.com}}
    \orcid{0000-0002-5607-9948}
    \affiliation{%
  \institution{\href{https://massive.com/}{Massive.com}}
  \city{Dublin}
  \country{Ireland}
  }

\author{Joe Dursun}
\authornotemark[1]
\affiliation{%
  \institution{\href{https://massive.com/}{Massive.com}}
  \city{Atlanta, Georgia}
  \country{USA}}

\author{Jarrett Blankenship}
\affiliation{%
  \institution{\href{https://massive.com/}{Massive.com}}
  \city{Atlanta, Georgia}
  \country{USA}}

\author{Katie Adams}
\affiliation{%
  \institution{\href{https://massive.com/}{Massive.com}}
  \city{Atlanta, Georgia}
  \country{USA}}

\author{Quinton Pike}
\affiliation{%
  \institution{\href{https://massive.com/}{Massive.com}}
  \city{Atlanta, Georgia}
  \country{USA}}

\renewcommand{\shortauthors}{Dolphin et al.}

\begin{abstract}
    We present a methodology for extracting structured risk factors from
  corporate 10-K filings while maintaining adherence to a predefined
  hierarchical taxonomy. Our three-stage pipeline combines LLM extraction
  with supporting quotes, embedding-based semantic mapping to taxonomy
  categories, and LLM-as-a-judge validation that filters spurious
  assignments. To evaluate our approach, we extract 10,688 risk factors from S\&P 500
  companies and examine risk profile similarity across industry clusters.
  Beyond extraction, we introduce autonomous taxonomy maintenance where an
  AI agent analyzes evaluation feedback to identify problematic categories,
  diagnose failure patterns, and propose refinements, achieving 104.7\%
  improvement in embedding separation in a case study. External validation
  confirms the taxonomy captures economically meaningful structure:
  same-industry companies exhibit 63\% higher risk profile similarity than
  cross-industry pairs (Cohen's $d=1.06$, AUC 0.82, $p<0.001$). The
  methodology generalizes to any domain requiring taxonomy-aligned
  extraction from unstructured text, with autonomous improvement enabling
  continuous quality maintenance and enhancement as systems process more documents.
\end{abstract}




\maketitle

\section{Introduction}

Corporate risk disclosure in annual reports (10-K filings) provides insightful commentary on the uncertainties facing public companies. Item 1A of these filings, the ``Risk Factors'' section, contains detailed discussions of material risks written in natural language by company management and legal teams. For investors, analysts, and risk managers, systematically analyzing these disclosures across hundreds or thousands of companies could reveal sector-wide risk patterns, emerging threats, and company-specific vulnerabilities. However, the unstructured nature of these documents and their substantial length (often spanning dozens of pages per company) makes manual systematic analysis at scale impractical.

A natural solution is to use large language models (LLMs) to extract structured risk information from these documents. Modern LLMs can indeed process 10-K risk sections and generate coherent lists of identified risks. However, this na\"ive approach creates a fundamental problem: without constraints, each LLM extraction produces idiosyncratic risk labels that may differ in terminology, granularity, and framing, even when describing the same underlying risk. One company might be tagged with ``currency fluctuation risk'' while another receives ``foreign exchange exposure,'' making systematic comparison impossible. The result is unstructured tags that defeat the purpose of automation---we trade one form of unstructured data for another.

Producing actionable data requires a method that maps LLM-extracted risks to a predefined, curated taxonomy. This allows consistent risk categorization across thousands of filings while preserving the flexibility of LLMs to identify diverse risk expressions in natural language. The challenge lies in the mapping itself: how do we reliably connect free-form risk descriptions extracted by an LLM to fixed categories in a hierarchical taxonomy, while ensuring that only genuinely relevant mappings are retained?

We propose a methodology that combines semantic embeddings with LLM-based validation to achieve robust taxonomy-aligned risk extraction. Our approach operates in three stages. First, we use an LLM with structured output to extract comprehensive risk factors and supporting quotes from the raw text. Second, we map these extracted risks to our predefined taxonomy using embedding-based similarity between the extracted quote and taxonomy category descriptions. Third, we employ an LLM-as-a-judge to validate each mapping, scoring how well the supporting quote actually matches the assigned taxonomy category and filtering out weak or spurious mappings. This validation stage serves dual purposes: it filters spurious mappings in production while generating systematic feedback for continuous taxonomy improvement. This multi-stage pipeline ensures that extracted risks adhere to a consistent taxonomy while maintaining high precision in their assignments.

Our contributions are:

\paragraph{A robust methodology for taxonomy-aligned risk extraction.} We present a complete pipeline that combines LLM extraction, embedding-based similarity mapping, and LLM validation to reliably map free-form risk descriptions to predefined taxonomy categories. The embedding-based mapping provides computational efficiency and semantic grounding, while the LLM-as-a-judge validation prevents false positives that would arise from na\"ive nearest-neighbor assignment alone.

\paragraph{A three-tier risk factors taxonomy for investment analysis.} We introduce a hierarchical taxonomy specifically designed for risk analysis in the investment context, with primary, secondary, and tertiary categories that organize risks at appropriate granularity for investment decision-making. Unlike existing taxonomies such as the Cambridge Taxonomy of Business Risks~\cite{cambridge2019taxonomy}, which subdivides categories like extreme weather into overly specific types (``Tropical Windstorm'' vs. ``Temperature Windstorm''), our taxonomy balances comprehensiveness with practical utility for financial analysis.

\paragraph{Autonomous taxonomy refinement workflow.} We present a systematic method for continuous taxonomy improvement using LLM-as-a-judge evaluation scores as feedback signals. An AI agent autonomously identifies problematic taxonomy categories by analyzing patterns in low-quality mappings, diagnoses root causes through examination of evaluation reasoning, generates candidate description refinements, and validates improvements using embedding-based testing that directly mirrors production matching. This creates a continuous improvement loop where taxonomy quality increases as the system processes more documents, reducing the human expertise burden in taxonomy maintenance. A case study demonstrates 104.7\% improvement in embedding separation for a pharmaceutical approval category through this autonomous workflow, transitioning taxonomy development from a static design problem to an iteratively improving system.

\paragraph{Empirical validation on S\&P 500 companies.} We evaluate our approach on 2024 10-K filings for S\&P 500 companies, extracting 10,688 validated risk factors mapped to our taxonomy. Industry clustering analysis demonstrates that companies in the same industry exhibit significantly more similar risk profiles than cross-industry pairs (63\% relative improvement in similarity, Cohen's $d=1.06$, $p<0.001$ across four statistical tests). This clustering emerges despite the taxonomy mapping having no access to industry codes, validating that extracted categories capture genuine economic risk dimensions. Finer industry granularity strengthens the signal (AUC improves from 0.733 for broad sectors to 0.822 for narrow industry definitions), and sector-specific analysis reveals intuitive patterns such as 83\% of banks tagged with interest rate risk versus 22\% of all companies. The production system has been deployed at Massive.com, processing companies across five years of historical filings and serving results via API.

\vspace{0.15cm}The methodology we present is not specific to risk factors or 10-K filings. It provides a general framework for extracting structured information from unstructured text while maintaining adherence to predefined categorical systems---a problem that arises in many domains where systematic analysis requires consistent labeling but source documents use unconstrained natural language.

\section{Related Work}

\paragraph{Financial document analysis and risk extraction.} The analysis of corporate disclosures using computational methods has been extensively studied. Early work focused on sentiment analysis of financial documents and identifying risk-related language through dictionary-based approaches~\cite{loughran2011liability}. More recent work has applied deep learning to classification and prediction based on financial text~\cite{du2025financial,rawte2018analysis,jun2024predicting}. Other work has explored using pre-trained language models fine-tuned on financial corpora~\cite{araci2019finbert,yang2020finbert,iacovides2024finllama}.
However, existing work primarily focuses on document-level sentiment analysis and broad classification tasks rather than structured extraction of specific risk factors aligned with pre-defined taxonomies.

\paragraph{Risk taxonomies for corporate analysis.} Several taxonomies have been proposed for classifying business risks. The Cambridge Taxonomy of Business Risks~\cite{cambridge2019taxonomy} provides a comprehensive framework but is designed for general risk management rather than investment analysis, leading to excessive granularity in some categories (e.g., subdividing weather risks by storm type). Industry-specific taxonomies exist (see, for example, \citet{kaufman2000banking} for banking and \citet{rabitti2025taxonomy} for cyber) but lack the breadth needed for cross-sector equity analysis. Our work contributes a taxonomy specifically designed for the investment use case, balancing coverage with practical granularity.

\paragraph{Structured information extraction with LLMs.} Recent advances in LLMs and the surrounding ecosystem have enabled reliable structured output through prompt engineering and function calling~\cite{pydantic_ai,anthropic_claude}.  While this solves the problem of output formatting, it does not address the challenge of mapping free-form LLM outputs to predefined categorical systems. Wei et al.~\cite{wei2022chain} showed that chain-of-thought prompting improves reasoning in complex tasks, and our extraction prompts follow this pattern by asking the LLM to identify risks while providing supporting quotes and reasoning---though this alone is insufficient to ensure taxonomy alignment. Examples of structured information extraction from unstructured text in the financial domain include~\cite{dolphin2024extracting,li2025extracting}.

\paragraph{Semantic similarity and embedding-based retrieval.} Embedding models that map text to dense vector representations have become central to semantic search and retrieval tasks~\cite{reimers2019sentence}. Recent work has shown that task-specific instructions can improve embedding quality for downstream applications~\cite{qwen3embedding,wang2023improving}. We leverage this capability using instruction-tuned embeddings to represent both taxonomy descriptions and extracted risk supporting quotes in a shared semantic space. The nearest-neighbor retrieval approach is standard in embedding-based semantic search~\cite{mikolov2013efficient}, but applying it to taxonomy mapping introduces the challenge of preventing spurious mappings when extracted content has no appropriate taxonomy category.

\paragraph{LLM-as-judge evaluation.} Using LLMs to evaluate outputs from other systems has emerged as a practical approach for tasks where traditional metrics are insufficient~\cite{zheng2023judging,dubois2023alpacafarm}. LLM judges have been applied to evaluate open-ended generation quality, factual consistency, and instruction following. Critically, recent work has shown that LLM judges can achieve high agreement with human evaluators when provided with clear scoring rubrics~\cite{zheng2023judging,liu2023g}. We extend this paradigm to the taxonomy mapping problem, using an LLM judge not just for post-hoc evaluation but as an integral component of the extraction pipeline to filter low-quality mappings in production.

\paragraph{Hybrid approaches to information extraction.} Our work sits in the broader context of hybrid systems that combine multiple techniques for robust information extraction. Recent systems have combined retrieval with generation~\cite{lewis2020retrieval}, used multiple models in ensemble~\cite{jiang2023llm}, or employed multi-stage pipelines with different models for different subtasks~\cite{khattab2022demonstrate}. Our three-stage approach (LLM extraction → embedding mapping → LLM validation) follows this philosophy, using each component where it excels: LLMs for understanding nuanced text and making holistic judgments, embeddings for efficient large-scale semantic similarity.

\section{Methodology}

\begin{figure*}[t!]
\centering
\begin{tikzpicture}[
    node distance=1cm and 2.5cm,
    stage/.style={rectangle, draw, rounded corners, minimum height=2.8cm, minimum width=3.2cm, align=center, fill=gray!8, font=\normalsize, thick},
    arrow/.style={->, >=stealth, very thick},
    feedback/.style={->, >=stealth, thick, dashed, gray},
]

\node[stage] (stage1) {
    \textbf{Stage 1: Extraction}\\[0.3cm]
    \small LLM extracts risk\\
    \small factors with supporting\\
    \small quotes from 10-K text
};

\node[stage, right=of stage1] (stage2) {
    \textbf{Stage 2: Mapping}\\[0.3cm]
    \small Embed quotes and\\
    \small match to pre-computed\\
    \small taxonomy embeddings
};

\node[stage, right=of stage2] (stage3) {
    \textbf{Stage 3: Validation}\\[0.3cm]
    \small LLM judge scores\\
    \small each mapping; filter\\
    \small low-quality matches
};

\draw[arrow] (stage1) -- (stage2);
\draw[arrow] (stage2) -- (stage3);

\draw[feedback] (stage3.south) -- ++(0,-0.6) -| node[pos=0.25, below, font=\small] {Low-quality mappings inform taxonomy refinement} (stage2.south);

\end{tikzpicture}
\caption{Three-stage pipeline for taxonomy-aligned risk extraction. Stage 1 uses an LLM to extract risk factors with supporting quotes from raw text. Stage 2 maps extracted risks to taxonomy categories using embedding-based semantic similarity. Stage 3 employs an LLM judge to validate mappings and filter spurious assignments, while low-quality mappings provide feedback for continuous taxonomy improvement.}
\label{fig:pipeline}
\end{figure*}
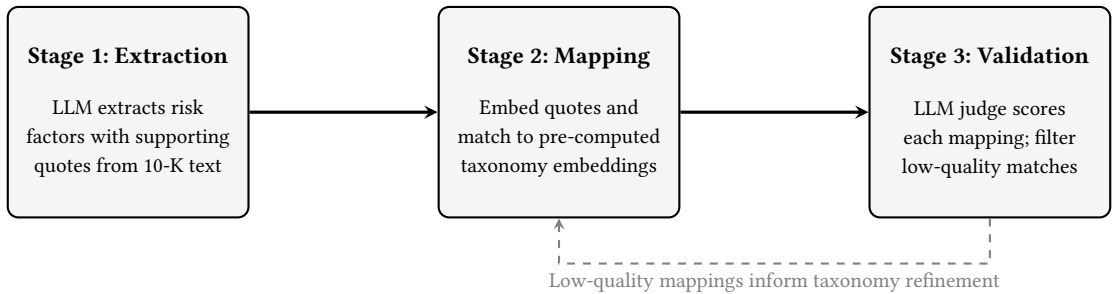

We present a three-stage pipeline for extracting risk factors from 10-K filings and mapping them to a predefined taxonomy (Figure~\ref{fig:pipeline}). The key insight is that no single technique suffices: LLMs excel at understanding nuanced text but produce inconsistent labels, embeddings enable efficient semantic matching but can produce spurious mappings, and validation is needed to ensure quality. Our approach combines these complementary strengths into a robust extraction system.

\subsection{Problem Formulation}

Let $\mathcal{T} = \{t_1, \ldots, t_N\}$ represent our risk factor taxonomy, where each category $t_i$ consists of a hierarchical label (primary, secondary, tertiary) and a natural language description $d_i$ that explains what risks belong in that category. The hierarchy allows primary and secondary labels to repeat across branches, while tertiary categories provide unique leaf-level classification. Given raw text $x$ from a 10-K Item 1A section, our goal is to produce a set of tuples $\{(t_{i_1}, q_1), \ldots, (t_{i_k}, q_k)\}$ where each $t_{i_j} \in \mathcal{T}$ is a taxonomy category and $q_j$ is a supporting quote from $x$ that justifies the assignment. Critically, we want high precision---only including taxonomy categories when genuinely supported by the text.

\subsection{Three-Tier Risk Taxonomy}

Our taxonomy organizes corporate risks into a three-tier hierarchy designed for investment analysis. The structure comprises seven primary categories (Strategic \& Competitive, Operational \& Execution, Financial \& Market, Technology \& Information, Regulatory \& Compliance, External \& Systemic, Governance \& Stakeholder), which subdivide into 28 secondary categories and further into 140 tertiary categories. Each tertiary category represents a specific risk type with an associated natural language description.

For example, under the primary category ``Financial \& Market,'' the secondary category ``Market \& Investment'' contains the tertiary risk ``Interest rate and yield curve risk,'' described as: \textit{``Risk from changes in interest rates or yield curves that can affect borrowing costs, investment returns, asset values, and overall financial performance. This includes risks from both rising and falling interest rate environments and their impact on various aspects of the business.''}

The taxonomy balances comprehensiveness with practical granularity. Unlike academic risk taxonomies that can overly subdivide non-consequential categories, our design groups related risks at a level useful for investment decision-making. Weather-related risks, for instance, are captured in a single category rather than being fragmented by storm type or temperature range.

\subsection{Stage 1: LLM-Based Risk Extraction}

The first stage uses an LLM to process the Item 1A text and extract a comprehensive list of risk factors without any constraint to match our taxonomy. We use Claude 4.5 Sonnet with structured output via function calling, configured with a task specific system prompt.

The LLM receives the full Item 1A text and must produce a list of risk exposures, where each exposure includes:
\begin{itemize}
    \item A free-form tag describing the risk
    \item A supporting quote from the original text justifying that risk's inclusion
\end{itemize}

This design follows chain-of-thought prompting principles. By requiring the LLM to provide supporting quotes, we encourage more grounded extraction and simultaneously obtain the text snippets needed for subsequent embedding-based matching. The structured output constraint ensures reliable parsing, with automatic retry on malformed responses.

The LLM is \textit{not} provided our taxonomy at this stage. Direct taxonomy matching proves unreliable due to context bloat, increasing costs and degrading performance. Instead, we separate comprehension from categorization.

\subsection{Stage 2: Embedding-Based Taxonomy Mapping}

Given the LLM-extracted risks with supporting quotes, we map each to the most semantically similar taxonomy category using dense embeddings. This approach provides computational efficiency (matching tens of extracted risks against 140 categories) while leveraging semantic similarity to find appropriate mappings.

\paragraph{Embedding generation.} We use the Qwen3 Embedding 0.6B~\cite{qwen3embedding} model, which supports task specific instructions that improve application specific embedding quality~\cite{qwen3embedding,wang2023improving}. We leverage this capability by prepending a task instruction to all text before embedding: \textit{``Classify risk factor text from an annual report into the most appropriate taxonomy category.''}

Taxonomy embeddings are computed once offline by embedding each category's natural language description with this instruction. At runtime, we embed each supporting quote from the LLM-extracted risks using the same instruction, ensuring both taxonomy descriptions and extracted quotes are represented in a shared, task-aligned latent space. The model is run locally, avoiding API costs and latency compared to cloud-based embedding services.

\paragraph{Nearest neighbor matching.} For each extracted risk with embedding $\mathbf{e}_q \in \mathbb{R}^{1024}$, we compute similarity to all taxonomy category embeddings $\{\mathbf{e}_{t_1}, \ldots, \mathbf{e}_{t_N}\}$ using dot product (equivalent to cosine similarity for normalized embeddings):
\begin{equation}
s(q, t_i) = \mathbf{e}_q^T \mathbf{e}_{t_i}
\end{equation}

We assign each extracted risk to its highest-similarity taxonomy category:
\begin{equation}
t^* = \argmax_{t_i \in \mathcal{T}}~s(q, t_i)
\end{equation}

This matching is implemented efficiently using matrix multiplication, computing all pairwise similarities in a single operation.

\paragraph{The spurious mapping problem.} Nearest neighbor assignment guarantees every extracted risk maps to \textit{some} taxonomy category, even when no appropriate category exists. Embeddings-based semantic search isn't a silver bullet. An LLM might extract a risk about ``regulatory approval delays for pharmaceutical products,'' which could map to a generic regulatory risk category despite our taxonomy having a more specific ``Clinical trial and regulatory approval risks'' category that would be a better fit---or worse, an extracted risk could be mapped to an entirely inappropriate category if the true risk isn't covered by our taxonomy at all. Heuristics based on the embedding similarity score alone are insufficient to detect these problematic mappings.

\subsection{Stage 3: LLM-as-Judge Validation}

To address spurious mappings, we employ an LLM as a judge~\cite{zheng2023judging} to evaluate each proposed mapping and assign a quality score. We initialize a separate LLM instance with a system prompt specifically designed for evaluation, distinct from the extraction stage. This validation step acts as a precision filter, discarding low-confidence assignments while retaining high-quality matches.

For each mapping of supporting quote $q$ to taxonomy category $t$ (with description $d$), we prompt the LLM with instructions to rate the match quality on a 1-5 scale, along with relevant context. The rating scale is defined as:
\begin{itemize}
    \item 5 = Excellent fit: Perfect match between text and classification
    \item 4 = Good fit: Appropriate with only minor issues
    \item 3 = Adequate fit: Reasonable but some gaps
    \item 2 = Poor fit: Significant misalignment
    \item 1 = Very poor fit: Clearly wrong classification
\end{itemize}

The LLM must provide both a numerical score and concise reasoning (one sentence) explaining the rating. We use structured output to ensure reliable parsing of the evaluation results. Due to the inherent latency of LLMs, evaluating each mapping sequentially would create a bottleneck, and so, we implement concurrent evaluation while respecting API rate limits.

\begin{table}
\centering
\caption{Distribution of LLM-as-judge quality scores across all evaluated mappings for S\&P 500 companies before quality filtering.}
\label{tab:quality_distribution}
\begin{tabular}{lcc}
\toprule
Quality Score & Count & Percentage \\
\midrule
1 (Very poor) & 1,291 & 7.8\% \\
2 (Poor) & 4,037 & 24.3\% \\
3 (Adequate) & 599 & 3.6\% \\
4 (Good) & 2,344 & 14.1\% \\
5 (Excellent) & 8,344 & 50.2\% \\
\midrule
Total & 16,615 & 100\% \\
\bottomrule
\end{tabular}
\end{table}

Table~\ref{tab:quality_distribution} shows the distribution of ratings. We retain only mappings with quality scores $\geq 4$, ensuring high precision in the final results. Lower-scored mappings are logged with their reasoning for taxonomy improvement---repeated low scores for similar extracted risks can indicate taxonomy gaps or categories needing refinement.

This logging serves dual purposes: filtering spurious mappings in production and providing systematic signals for taxonomy improvement during development. When a category accumulates many low-quality mappings with similar patterns in the reasoning (e.g., ``text discusses EU regulatory approval but description emphasizes FDA''), this indicates the category description needs refinement to better capture the intended scope or distinguish true positives from common false positives. Section~\ref{sec:taxonomy_refinement} demonstrates this feedback loop through a detailed case study of autonomous taxonomy refinement using LLM evaluation outputs.

\subsection{Deduplication and Final Output}

The pipeline may produce multiple LLM-extracted risks that map to the same taxonomy category (e.g., different phrasings of cybersecurity risk). After quality filtering, we deduplicate by taxonomy category, retaining the highest-quality mapping for each category. This ensures each taxonomy category appears at most once in the final output.

The final output for a given 10-K filing is an object where each element represents one risk factor:
\begin{itemize}
    \item Primary, secondary, and tertiary category labels
    \item Supporting quote from the original Item 1A text
    \item Original LLM-generated tag (for interpretability)
    \item Quality score (for downstream filtering or analysis)
\end{itemize}

\section{Evaluation}

We evaluate our methodology on 2024 10-K filings for S\&P 500 companies, addressing two distinct questions. First, can we systematically improve taxonomy quality using LLM-as-a-judge feedback signals to refine problematic category descriptions? Second, do the extracted risk profiles capture economically meaningful structure---specifically, whether companies in the same industry exhibit more similar risk profiles than cross-industry pairs? The first question tests whether the methodology enables continuous improvement of the taxonomy itself, while the second tests whether the extracted risks reflect genuine sectoral patterns rather than arbitrary categorizations. We find that autonomous taxonomy refinement yields substantial improvements in embedding separation (104.7\% for a pharmaceutical approval category), while industry clustering analysis demonstrates strong and statistically significant sectoral patterns (63\% higher similarity for same-industry pairs, Cohen's $d=1.06$, $p<0.001$).

\subsection{Data Acquisition}

We obtain risk factor sections (Item 1A) from 10-K filings using Massive.com's SEC AI-Ready APIs\footnote{\href{https://massive.com/docs/rest/stocks/filings/10-k-sections}{https://massive.com/docs/rest/stocks/filings/10-k-sections}}. These APIs provide direct access to parsed Item 1A sections, eliminating the need for custom XBRL parsing. For a given ticker symbol and fiscal year, the API returns clean plain text of the risk factors section, typically ranging from 4,000 to 20,000 words. This pre-processing is helpful because raw SEC filings contain inconsistent formatting, nested tags, and presentation markup that complicate downstream analysis.

\subsection{Taxonomy Quality Analysis and Iterative Refinement}
\label{sec:taxonomy_refinement}

The LLM-as-judge evaluation scores provide more than a filtering mechanism---they offer systematic feedback for improving the taxonomy itself. When a category consistently receives low-quality scores across many mappings, this signals potential issues with the category description that can be addressed through iterative refinement. We demonstrate this through a detailed case study of an autonomous AI agent workflow that identifies problematic categories, analyzes failure patterns, proposes refinements, and validates improvements through embedding-based testing.

\paragraph{Quality score distribution.} Across all 500 S\&P 500 companies, the LLM-as-judge evaluated 16,615 candidate mappings from the embedding-based matching stage. Again, see Table~\ref{tab:quality_distribution} for the distribution of evaluation scores. While the majority of mappings (50.2\%) received the highest quality score of 5, a substantial fraction (32.1\%) received scores of 1 or 2, indicating poor matches between the extracted risk text and the assigned taxonomy category. These low-quality mappings are filtered from production results but logged with their evaluation reasoning for taxonomy analysis.

\paragraph{Autonomous identification of problematic categories.} An AI agent with custom tools analyzed the logged low-quality mappings to identify taxonomy categories requiring attention. By aggregating mappings with scores $< 4$ by category, the agent identified ``Pharmaceutical and medical device approval'' as accumulating 179 low-quality mappings---one of the highest among all 140 tertiary categories. The quality score distribution for this category was notably worse than the overall distribution: 78.1\% of its mappings received scores of 1 or 2, compared to 32.1\% across all categories (Table~\ref{tab:quality_distribution}).

The category's original description was: \textit{``Risk for healthcare companies from regulatory approval processes, including FDA approvals, market access restrictions, or changes in regulatory requirements that could delay product launches, limit market opportunities, or require additional studies.''} The agent retrieved all 179 low-quality mappings along with their LLM evaluation reasoning to analyze failure patterns.

\paragraph{Pattern analysis and root cause identification.} The agent systematically analyzed the evaluation reasoning across the low-quality mappings, identifying four major failure patterns:

\begin{enumerate}
\item \textit{Geographic scope confusion} (30\% of cases): While the description mentioned ``including FDA approvals,'' many extracted risks specifically discussed EU Medical Devices Regulation, UK UKCA marking, or EMA processes. These were legitimate regulatory approval risks but received low scores with reasoning like ``text specifically discusses EU/EMA regulations, not FDA approval processes.'' The phrasing ``including FDA'' inadvertently suggested FDA primacy rather than multi-jurisdictional coverage.

\item \textit{Post-market vs. pre-market misalignment} (20\%): The phrase ``market access restrictions'' attracted post-approval compliance risks (off-label promotion penalties, Phase 4 trial requirements) that should not match this category. Evaluation reasoning correctly noted these describe post-market enforcement rather than pre-market approval processes.

\item \textit{Reimbursement conflation} (15\%): ``Market access restrictions'' also incorrectly attracted reimbursement-related risks such as physician adoption barriers due to unclear payor policies. These belong in the separate ``Reimbursement and pricing pressure'' category.

\item \textit{Non-healthcare regulatory approvals} (10\%): Generic mentions of ``approval processes'' incorrectly attracted risks about FCC regulations, DOE export permits, and M\&A regulatory clearance, which the LLM judge correctly rejected as irrelevant to pharmaceutical/device approval.
\end{enumerate}

\paragraph{Automated description refinement and testing.} Based on these patterns, the agent autonomously generated multiple candidate description refinements and implemented an embedding-based testing methodology to evaluate them. The agent created a test set of cases: true positives representing genuine regulatory approval risks across multiple jurisdictions (FDA, EMA, MHRA) and false positives drawn from the identified failure patterns.

For each candidate description, the agent: (1) generated embeddings using the same Qwen3-Embedding model with task instructions, (2) computed cosine similarity against all test cases, and (3) measured embedding separation quality as the difference between the average similarity scores of true positive test cases versus false positive. 
In some sense, this process can be thought of as analogous to contrastive learning: we are trying to find category descriptions that maximize similarity with true positive test cases while minimizing similarity with false positive cases.

Table~\ref{tab:description_variants_regulatory} shows selected variants tested by the agent. The key insight was that removing the ambiguous phrase ``market access restrictions'' while explicitly listing multiple health authorities (FDA, EMA, and other agencies) and emphasizing the pre-market approval stage would address the identified failure patterns.

\begin{table*}
\centering
\caption{Performance of description variants for ``Pharmaceutical and medical device approval'' tested by the autonomous agent. Separation measures the difference between average true positive and false positive similarity scores in the latent space (higher is better).}
\label{tab:description_variants_regulatory}
\small
\begin{tabular}{p{10.5cm}ccc}
\toprule
Description Variant & Avg TP & Avg FP & Separation \\
\midrule
\textbf{Original}: Risk for healthcare companies from regulatory approval processes, including FDA approvals, market access restrictions, or changes in regulatory requirements that... & 0.522 & 0.457 & 0.064 \\
\midrule
\textbf{V1}: Risk ... from regulatory approval processes, including FDA, EMA, and other health authority approvals, or changes in regulatory requirements that could delay product launches... & 0.524 & 0.436 & 0.089 \\
\midrule
\textbf{V2}: Risk from ... regulatory approval delays or rejections by health authorities (FDA, EMA, MHRA, and other agencies) due to clinical trial requirements, review process delays, or requests... & 0.556 & 0.434 & 0.122 \\
\midrule
\textbf{Final (V3)}: Risk from pre-market regulatory approval delays or rejections for pharmaceutical and medical device products by health authorities including FDA, EMA, and international agencies. This includes clinical trial failures, regulatory review delays, and requirements... & 0.564 & 0.433 & \textbf{0.132} \\
\bottomrule
\end{tabular}
\end{table*}

The final refined description achieved 0.132 separation, a \textit{104.7\% improvement} over the original's 0.064. Critical changes included: (1) removing ``market access restrictions'' which caused the post-market and reimbursement confusion, (2) explicitly listing ``FDA, EMA, and international agencies'' to signal multi-jurisdictional coverage rather than FDA primacy, (3) adding ``pre-market'' to distinguish from post-approval compliance, and (4) including specific regulatory approval terminology (``clinical trial failures, regulatory review delays'').

\paragraph{Implications for continuous taxonomy improvement.} This case study demonstrates an end-to-end autonomous workflow for taxonomy refinement: the AI agent identifies problematic categories from LLM-as-a-judge outputs, retrieves and analyzes low-quality mappings to diagnose root causes, generates candidate description refinements, implements quantitative testing methodology, and validates improvements before recommending deployment. The entire process---from identifying the ``Pharmaceutical and medical device approval'' category as problematic to producing the refined description with measured 104.7\% improvement in embedding separation---was performed autonomously, with human oversight but no manual intervention in the analysis or refinement steps.

This autonomous refinement capability enables continuous taxonomy improvement as the system processes new documents. As low-quality mappings accumulate for various categories, the agent can periodically identify categories needing attention and propose evidence-based refinements. The embedding-based testing methodology provides objective measurement of improvements, ensuring that refinements genuinely enhance the semantic separation between true and false positives rather than merely changing the description arbitrarily. While the case study presented here focused on a single category, the methodology applies systematically across all taxonomy categories---the agent can identify and refine multiple problematic categories using the same workflow, with prioritization based on the volume and severity of low-quality mappings.

\subsection{Industry Clustering Validation}

The preceding evaluation demonstrates internal consistency through embedding separation metrics and autonomous taxonomy refinement, but does not validate whether the extracted taxonomy captures economically meaningful structure. We now test whether companies in the same industry exhibit similar risk profiles when similarity is computed purely from taxonomy assignments with no industry information in the matching process.

\subsubsection{Experimental Design}

We construct a binary risk matrix $\mathbf{R} \in \{0,1\}^{n \times k}$ where $n=500$ companies and $k=137$ represents the tertiary taxonomy categories appearing in our S\&P 500 sample\footnote{Three categories were not assigned to any of the sample companies, accounting for the discrepancy between $k=137$ and the $140$ total tertiary categories in the taxonomy.}, with $R_{ij}=1$ if company $i$ was tagged with risk category $j$.

Not all risks are equally informative for distinguishing industries. Generic risks like ``data breaches'' appear in 270 companies (58\%), providing little discriminative signal---most companies face cybersecurity concerns regardless of industry. Specialized risks like ``clinical trial risks'' appear in only 14 companies (3\%), strongly indicating pharmaceutical firms. We therefore weight each risk category inversely to its prevalence, upweighting rare risks that distinguish industries while downweighting common risks shared across sectors.

This approach is inspired by inverse document frequency (IDF) weighting from information retrieval, adapted to our binary setting. Since companies either have a risk or not (no frequency counts), the term frequency component is always 1, leaving only the inverse prevalence weighting. For each category $j$, we compute:
\begin{equation}
w_j = \log \frac{n}{\sum_i R_{ij} + 1}
\end{equation}
and weight the matrix as $\tilde{R}_{ij} = R_{ij} \cdot w_j$. The logarithm creates balanced weights: very rare risks appearing in 2 companies receive weight $\log(500/2) = 5.52$, while common risks in 250 companies receive weight $\log(500/250) = 0.69$, yielding an 8-fold range. Linear inverse weighting, without the log transformation, produces extreme ratios (250 versus 2, a 125-fold range) that allow ultra-rare features to dominate similarity, degrading classification performance.

We compute pairwise company similarity using cosine distance on weighted risk vectors and test whether same-industry pairs (sharing 2-digit SIC codes) exhibit higher similarity than different-industry pairs. This provides external validation: the taxonomy mapping has no access to industry information, yet if it captures genuine business risks, industry structure should emerge naturally from the risk profiles.

\subsubsection{Results} 
\begin{figure}
\centering
\includegraphics[width=1\linewidth]{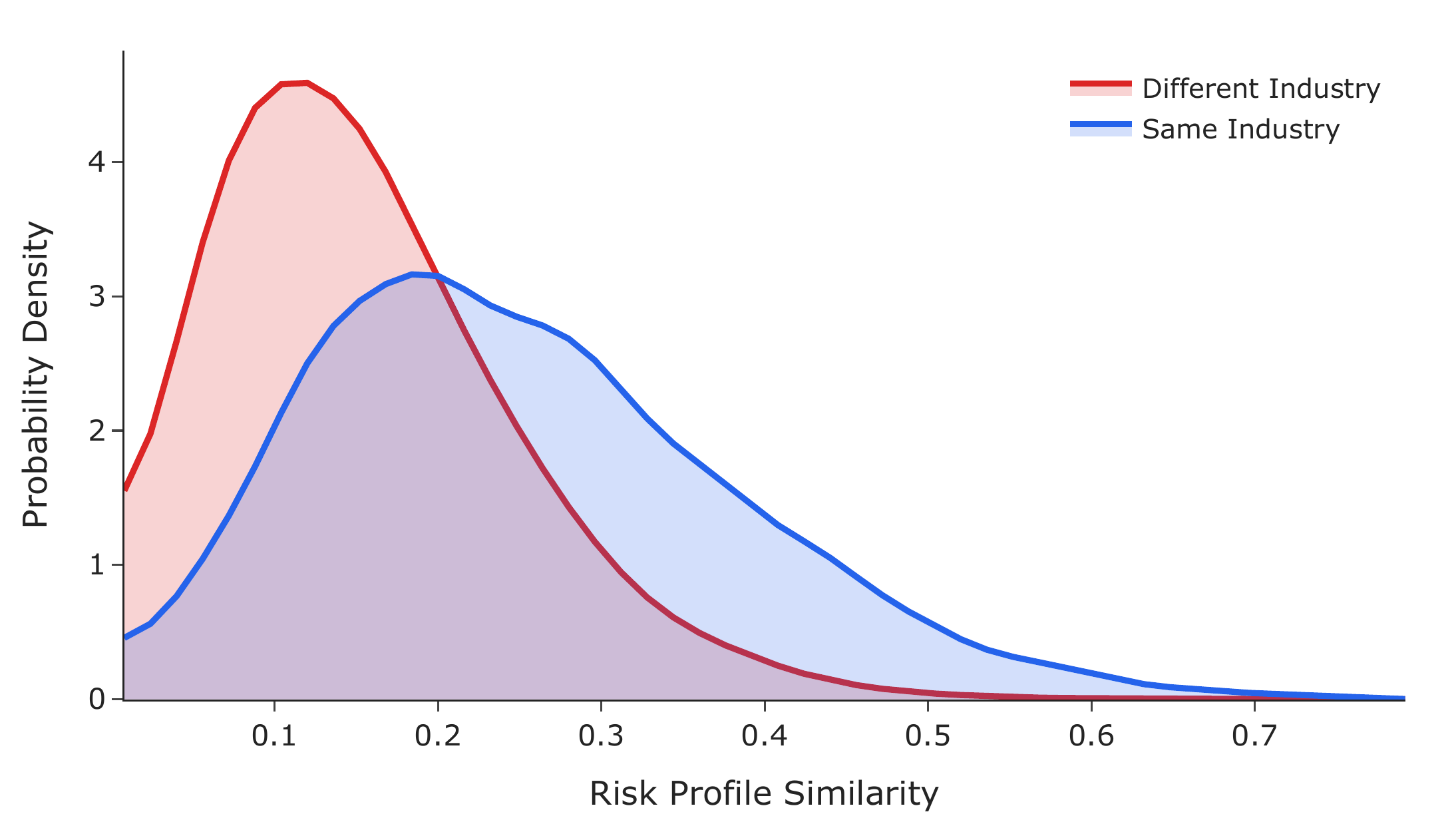}
\caption{Risk profile similarity distributions for same-industry versus different-industry company pairs using weighted similarity. Companies sharing 2-digit SIC codes (blue, 5,263 pairs) exhibit substantially higher similarity than companies in different industries (red, 101,228 pairs). The clear distributional separation demonstrates that extracted risk profiles naturally capture sectoral information despite no industry information in the taxonomy mapping process.}
\label{fig:similarity_distribution}
\end{figure}

Figure~\ref{fig:similarity_distribution} shows similarity distributions for same-industry versus different-industry pairs. Companies in the same industry exhibit substantially higher risk profile similarity (mean 0.254) compared to companies in different industries (mean 0.156), a 63\% relative increase. The distributions show clear separation: same-industry pairs form a distinct mode shifted rightward from the different-industry distribution.

Four statistical tests confirm this difference is highly significant and represents a large effect. Welch's $t$-test yields $t=54.59$ ($p<0.001$). The Kolmogorov-Smirnov test shows the entire distributions differ significantly ($D=0.341$, $p<0.001$), not just their central tendencies. A permutation test with 10,000 iterations confirms robustness to distributional assumptions ($p<0.001$). Cohen's $d=1.062$ indicates a very large effect size by conventional standards ($d>0.8$), demonstrating both statistical and practical significance.

\begin{figure}
\centering
\includegraphics[width=0.9\linewidth]{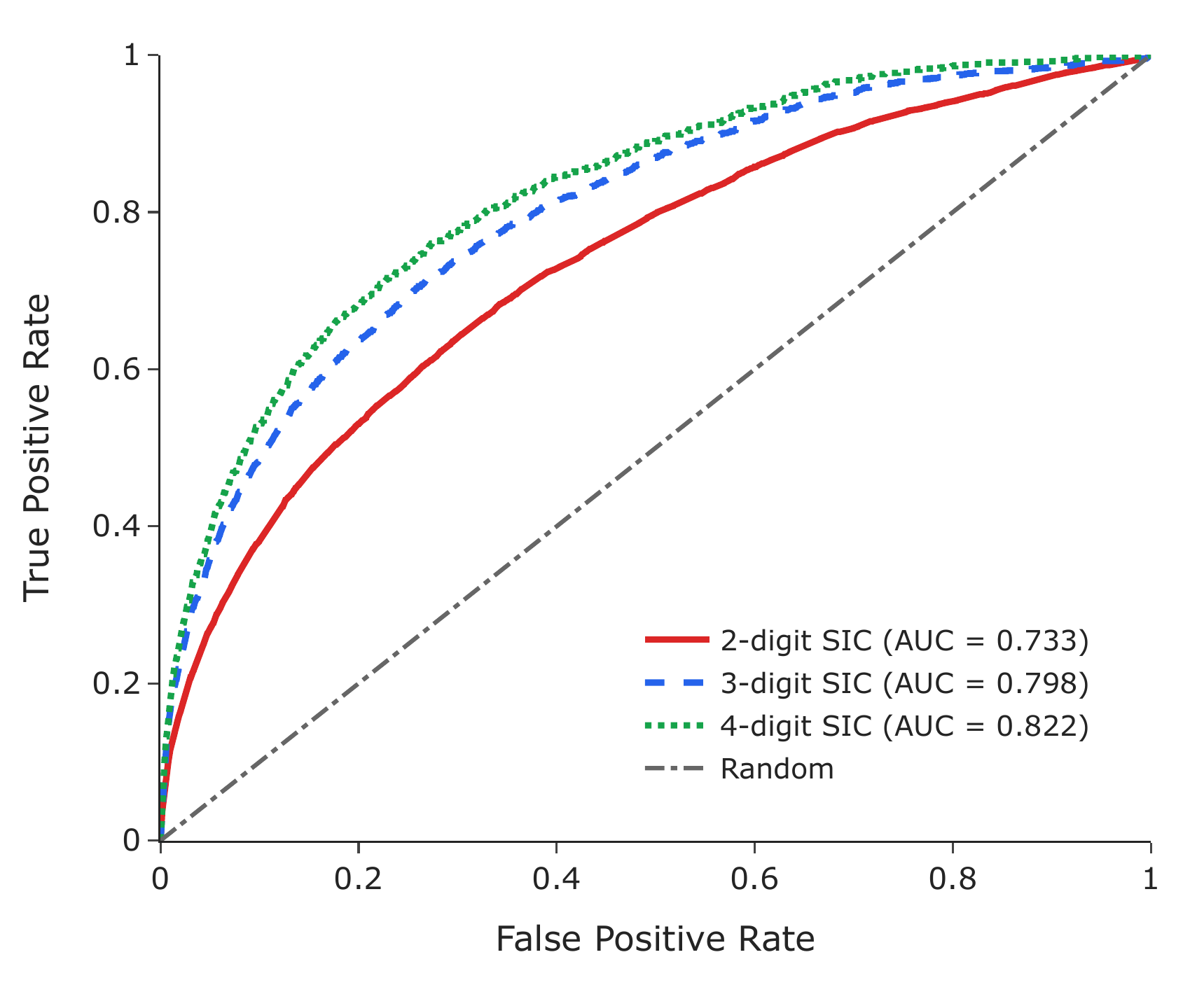}
\caption{ROC curves for predicting industry membership from weighted risk similarity across three SIC code granularities. Finer industry definitions (4-digit SIC, AUC = 0.822) produce stronger clustering than broad sectors (2-digit, AUC = 0.733), demonstrating that extracted risk profiles capture industry-specific patterns at multiple levels of specificity. All curves substantially outperform random classification (diagonal, AUC = 0.5).}
\label{fig:roc_curve}
\end{figure}

Figure~\ref{fig:roc_curve} tests whether risk similarity predicts industry membership by computing ROC curves for binary classification using similarity as the decision variable. That is, for each company pair, we predict whether they share industry membership based on their similarity. At 2-digit SIC granularity (broad sectors like "Depository Institutions" or "Chemicals"), AUC reaches 0.733, substantially above random classification (0.5). More granular industry codes strengthen the signal: 3-digit SIC achieves AUC 0.798, while 4-digit (highly specific industries) reaches 0.822. This progression demonstrates that our taxonomy captures fine-grained industry characteristics, not merely coarse sectoral patterns.

These results validate that our extraction pipeline captures economically meaningful risk dimensions rather than generic boilerplate. The taxonomy was designed for investment analysis without reference to industry codes, yet the risk profiles naturally capture sectoral information.

\subsubsection{Sensitivity to granularity} Table~\ref{tab:clustering_sensitivity} shows clustering strength across taxonomy levels and SIC digit precision. Taxonomy granularity proves critical: primary categories (7 total) produce essentially random classification (AUC 0.509), secondary categories (28) show moderate clustering (AUC 0.633), while tertiary categories (137) yield strong industry signal (AUC 0.733). This progression reflects that coarse categorization collapses industry-specific variation---all companies face "strategic and competitive" risks, but tertiary manifestations differ meaningfully by industry.

\begin{table}
\centering
\caption{Industry clustering validation across taxonomy and SIC granularities using weighted similarity. All differences significant at $p<0.001$ across all statistical tests.}
\label{tab:clustering_sensitivity}
\begin{tabular}{llcccc}
\toprule
Taxonomy & SIC & \multicolumn{2}{c}{Mean Similarity} & & \\
Level & Digits & Same Ind. & Diff. Ind. & $\Delta$ & AUC \\
\midrule
Primary   & 2 & 0.752 & 0.751 & 0.001 & 0.509 \\
Secondary & 2 & 0.476 & 0.380 & 0.096 & 0.633 \\
Tertiary & 2 & 0.254 & 0.156 & 0.098 & 0.733 \\
Tertiary & 3 & 0.292 & 0.158 & 0.134 & 0.798 \\
Tertiary & 4 & 0.307 & 0.159 & 0.148 & 0.822 \\
\bottomrule
\end{tabular}
\end{table}

Industry code precision amplifies clustering for tertiary categories: moving from 2-digit to 4-digit SIC increases AUC from 0.733 to 0.822, with absolute similarity differences growing from 0.098 to 0.148. Finer industry definitions (e.g., "pharmaceutical preparation" versus "chemicals and allied products") correspond to more specialized risk profiles that our taxonomy successfully distinguishes. Notably, different-industry similarity remains stable around 0.16 regardless of SIC granularity, while same-industry similarity increases from 0.254 (2-digit) to 0.307 (4-digit), confirming that narrow industry definitions genuinely share more similar risk profiles rather than the effect arising from statistical artifacts.

These results validate our design choices. Tertiary granularity captures industry-specific risk patterns that coarser levels collapse. The strong signal at 2-digit SIC (broad sectors) demonstrates the taxonomy captures fundamental industry characteristics, not narrow idiosyncrasies, providing a practical balance for cross-sector analysis.

\subsection{Sector-Specific Risk Profiles}

The preceding analysis demonstrates that risk profiles cluster by industry in aggregate, but what does this look like for a specific sector? We examine SIC 60 (Depository Institutions), containing 12 banks including Bank of America, JPMorgan Chase, and State Street.

\begin{figure}
\centering
\includegraphics[width=1\linewidth]{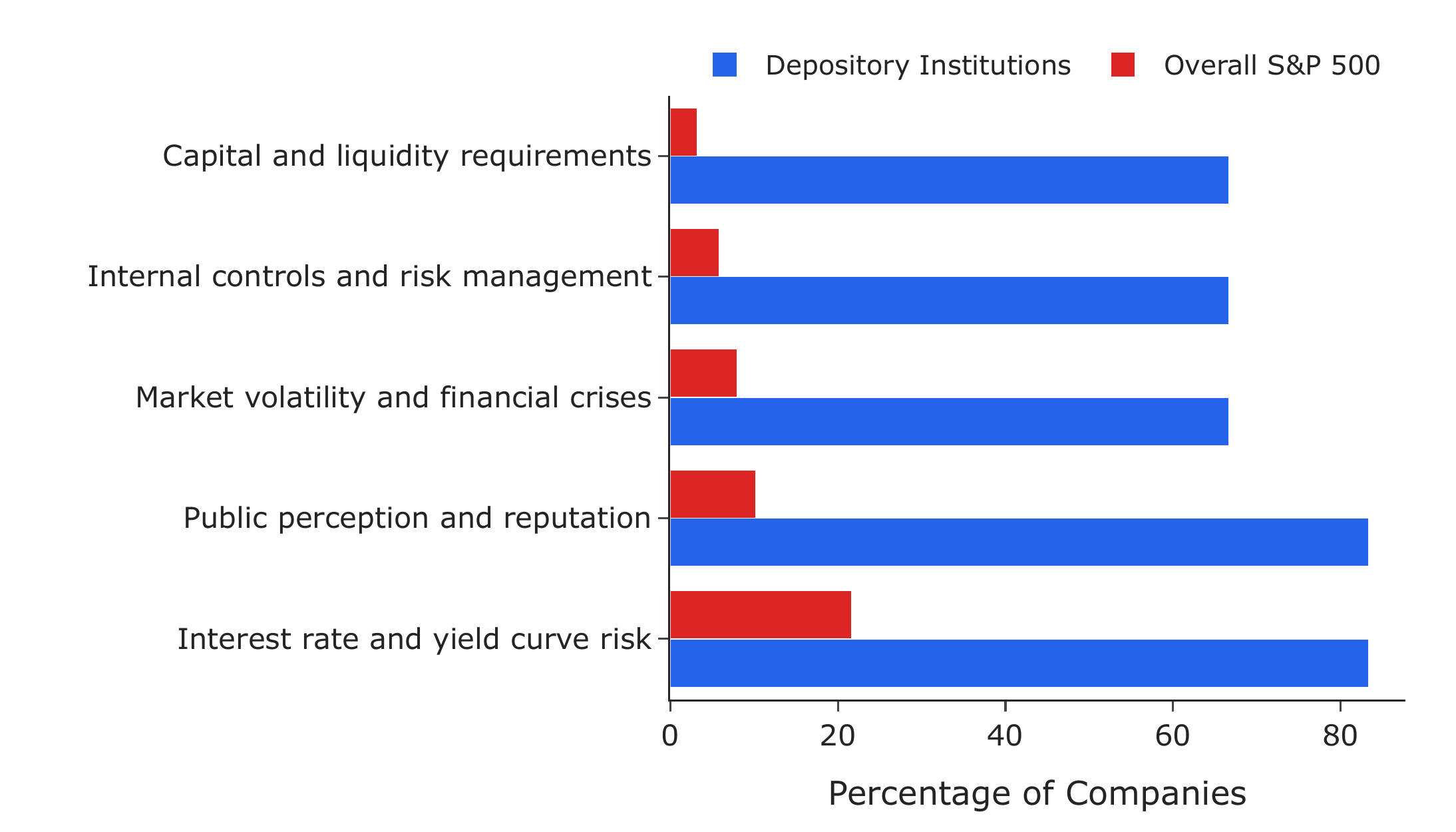}
\caption{Risk category prevalence for Depository Institutions (SIC 60, $n=12$) versus overall S\&P 500 population. Bars show the percentage of companies tagged with each risk. Banks exhibit systematic overrepresentation of finance-specific risks: 83\% have interest rate risk versus 22\% overall, 67\% have capital requirements versus 3\% overall. Risks are sorted by overall S\&P prevalence (highest at top) to emphasize sector-specific enrichment.}
\label{fig:sector_enrichment}
\end{figure}

Figure~\ref{fig:sector_enrichment} shows the most overrepresented risk categories for banks. Finance-specific risks appear with dramatically higher prevalence than in the overall S\&P 500 population: interest rate and yield curve risk appears in 83\% of banks versus only 22\% of all companies. Capital and liquidity requirements appears in 67\% of banks versus 3\% overall---this regulatory requirement is largely specific to financial institutions. Credit risk appears in 50\% versus 17\%, market volatility in 67\% versus 8\%, and public perception in 83\% versus 10\%. The latter reflects banks' particular sensitivity to reputational damage given depositor confidence and regulatory scrutiny.

Conversely, risks common in other sectors rarely appear for companies in the Depository Institutions SIC class. Raw material availability appears in only 8\% versus 42\% of all companies, and safety and environmental regulations in 8\% versus 45\% overall. This pattern---strong enrichment for sector-relevant risks, systematic depletion for irrelevant ones---demonstrates that our taxonomy captures what makes banks economically distinctive: interest rate sensitivity, regulatory capital requirements, credit quality, and systemic financial risk, while appropriately filtering out manufacturing and supply chain concerns that dominate other industries.

\section{Conclusion}

We present a methodology for extracting structured risk information from corporate disclosures while maintaining adherence to a predefined taxonomy. The core challenge---mapping free-form LLM-extracted text to fixed categories without producing spurious assignments---is addressed through a three-stage pipeline combining LLM extraction, embedding-based semantic matching, and LLM-as-a-judge validation. This hybrid approach leverages the complementary strengths of each component: LLMs understand nuanced language and make holistic judgments, embeddings provide efficient semantic similarity at scale, and the validation layer filters false positives that arise from nearest-neighbor assignment alone.

Our evaluation demonstrates the methodology's effectiveness on S\&P 500 companies' 2024 10-K filings, extracting 10,688 validated risk factors across the pre-defined taxonomy categories. Beyond validating extraction quality, we introduce autonomous taxonomy maintenance as a distinct contribution: an AI agent analyzes LLM-as-a-judge evaluation scores to systematically identify problematic categories, diagnose failure patterns, and propose evidence-based refinements. This creates a continuous improvement loop where taxonomy quality increases as the system processes more documents, reducing the human expertise burden in taxonomy design and maintenance. A case study demonstrates 104.7\% improvement in embedding separation through autonomous refinement of a pharmaceutical approval category.

External validation through industry clustering confirms the taxonomy captures economically meaningful risk dimensions. Companies in the same industry exhibit 63\% higher risk profile similarity than cross-industry pairs when measured using inverse prevalence weighted similarity on taxonomy assignments. This clustering emerges despite the mapping process having no access to industry codes, with strong statistical significance across four tests and high predictive power for industry membership (AUC 0.733-0.822 depending on industry granularity). Sector-level analysis reveals intuitive patterns: 83\% of banks are tagged with interest rate risk versus 22\% of all companies, while manufacturing concerns like raw material availability appear in only 8\% of banks versus 42\% overall.

The methodology we present extends beyond risk factor extraction from 10-K filings. It provides a general framework for mapping unstructured text to predefined categorical systems---a problem arising in many domains where systematic analysis requires consistent labeling but source documents use unconstrained natural language. The approach could be applied to other SEC filing sections (MD\&A, business descriptions), temporal analysis tracking risk evolution across years, or entirely different domains requiring taxonomy-aligned extraction from unstructured text.

Several limitations warrant discussion. LLM-based extraction and validation incur API costs and latency that may constrain real-time applications or extremely large-scale processing, though batch processing and caching mitigate these concerns in practice. Taxonomy design requires upfront domain expertise to define appropriate categories and descriptions, though our autonomous improvement workflow reduces ongoing maintenance burden as the system learns from evaluation feedback. The current implementation focuses on English-language corporate filings, though the methodology itself is language-agnostic given multilingual LLMs and embedding models.

The combination of robust extraction, autonomous improvement, and external validation provides an effective framework for taxonomy-aligned information extraction from unstructured text. As LLMs become more capable and embedding models more sophisticated, the core challenge shifts from raw extraction capability to ensuring consistency, precision, and alignment with predefined categorical systems. Our methodology addresses this challenge through hybrid techniques and continuous refinement, enabling systematic analysis at scale while maintaining the structured outputs that downstream applications require. The extracted data is available via the \href{https://massive.com/docs/rest/stocks/filings/risk-factors}{Risk Factors API at Massive.com}.


\bibliographystyle{ACM-Reference-Format}
\bibliography{sample-base}



\end{document}